# Using curvature to distinguish between surface reflections and vessel contents in computer vision based recognition of materials in transparent vessels

*Sagi Eppel*

Department of Materials Science and Engineering, Technion – Israel Institute of Technology, Haifa 32000, Israel.

E-mail: sagieppel@gmail.com
Tel: +972 523 202 516

## Abstract

The recognition of materials and objects inside transparent containers using computer vision has a wide range of applications, ranging from industrial bottles filling to the automation of chemistry laboratory. One of the main challenges in such recognition is the ability to distinguish between image features resulting from the vessel's surface and image features resulting from the material inside the vessel. Reflections and the functional parts of a vessel's surface can create strong edges that can be mistakenly identified as corresponding to the vessel contents, and cause recognition errors. The ability to evaluate whether a specific edge in an image stems from the vessel's surface or from its contents can considerably improve the ability to identify materials inside transparent vessels. This work will suggest a method for such evaluation, based on the following two assumptions:

1) Areas of high curvature on the vessel surface are likely to cause strong edges due to changes in reflectivity, as is the appearance of functional parts (e.g. corks or valves).

2) Most transparent vessels (bottles, glasses) have high symmetry (cylindrical). As a result the curvature angle of the vessel's surface at each point of the image is similar to the curvature angle of the contour line of the vessel in the same row in the image.

These assumptions, allow the identification of image regions with strong edges corresponding to the vessel surface reflections. Combining this method with existing image analysis methods for detecting materials inside transparent containers allows considerable improvement in accuracy.

# 1. Introduction

Computer vision-based identification of materials and objects inside transparent vessels is an important challenge[1-32], with applications ranging from chemistry laboratory automation[13,33,34] to industrial bottle filling.[12,22,35] Many types of materials such as liquids, powders and gels, are dealt with almost exclusively while carried in container vessels. The ability to deal with such material depends on the ability to accurately identify its position in a vessel.

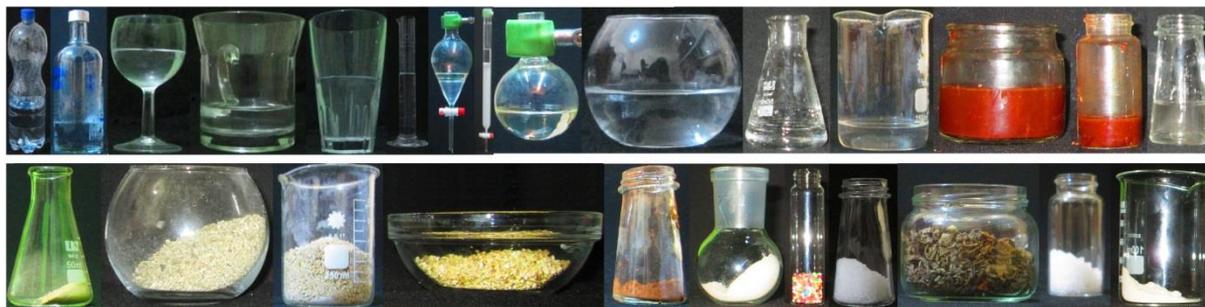

**Figure 1) Various materials in glassware vessels from both everyday life and the chemistry laboratory.**

A major problem in the computer-based recognition of materials and objects inside transparent containers is the identification of which edges and features in the image correspond to the surface of the vessel and which correspond to the material inside the vessel.[1,2] Edges in the image that result from the vessel's surface or functional parts (e.g. corks or valves) are often mistakenly attributed to materials inside the vessel, which can cause false recognition of the vessel's content.[1,2] This work suggests a general method to improve the recognition of objects and materials inside transparent vessels by distinguishing image features produced by the vessel's surface from those caused by the vessel's content. The general assumption is that areas of the vessel surface with sharp curvature angles, and a vessel's functional parts (corks, valves) are more likely to show strong edges due to reflectivity changes.[36-39] The greater the vessel surface curvature in a given area, the higher the probability of strong edges stemming from the surface in this area. This assumption is true for both Lambertian surfaces and smooth specular surfaces.[36-39] For this work 'curvature angle' will be loosely defined as the amount of change in a surface/line angle at a given point.[40] Evaluation of the vessel's surface curvature angle in the image is based on the assumption that the majority of transparent vessels (jars, bottles, cups) have high symmetry, which implies a similarity between the curvature of the vessel outline in the image and the curvature of every surface point in that row of the image. The curvature of the vessel surface at a specific point in the image is thus similar to the curvature of the vessel contour line in the same row of the image. This method was tested using images of various types of glassware containers from everyday life and the chemistry laboratory (e.g. bottles, beakers, flasks, jars, columns, vials and separation-funnels).[41] Combining this approach with an existing image analysis method for tracing liquid levels and the boundaries of materials inside transparent vessels, allows considerable improvement in recognition accuracy.

## 2. Method

Most vessels used for dealing with liquids and solids are either cylindrical (axisymmetric) or have some other high symmetry. This implies that the curvature of the surface of the vessel in a given point of an image is related to the curvature of the vessel contour line in that row of the image (Figure 2a). If the contour of the vessel in the image is known, then this method is a straightforward approach to identifying the curvature of the vessel surface in every point of the image.

### 2.1. Scale dependency of contour line curvature angle

While extracting the curvature angle of the vessel contour is relatively straightforward, there is still a problem regarding the scale at which the curvature angle is measured. In different scales, the curvature angle can be very different (Figure 2b). To account for this, the angle of the curvature at a given point was taken as the average of the curvature angle over a range of scales (Figure 2b-c). In this work the curvature angle at a given point was averaged over scale ranging between 1/50 and 1/14 of the vessel height in the image. Several further adjustments are possible, including smoothing the vessel contour line to reduce noise and normalising the weight of angles at large scales, however, it was found that none results in a clear improvement in recognition accuracies and they were therefore abandoned.

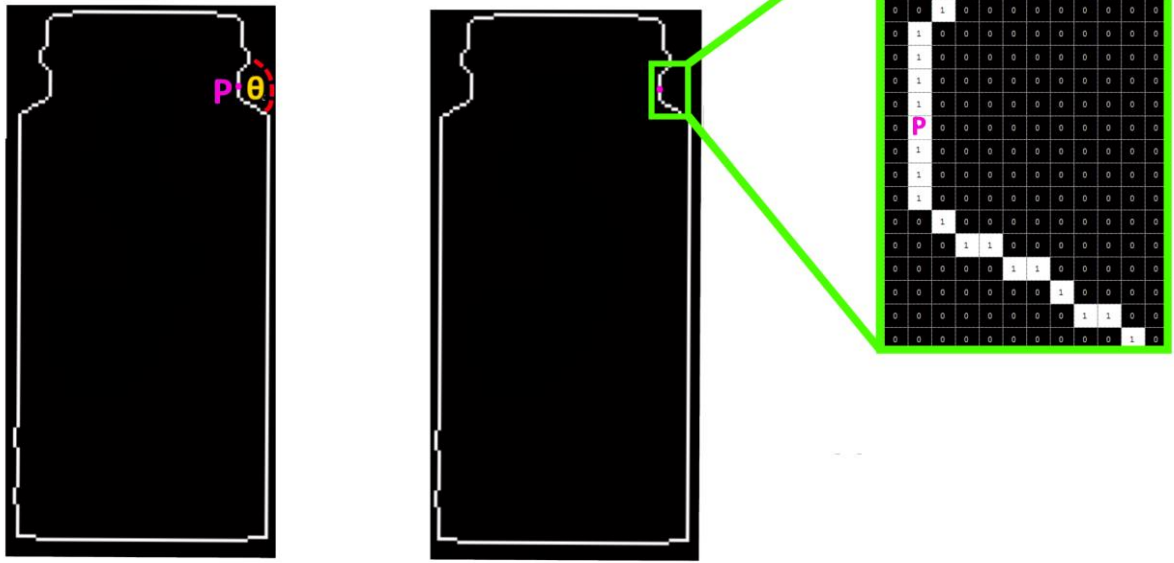

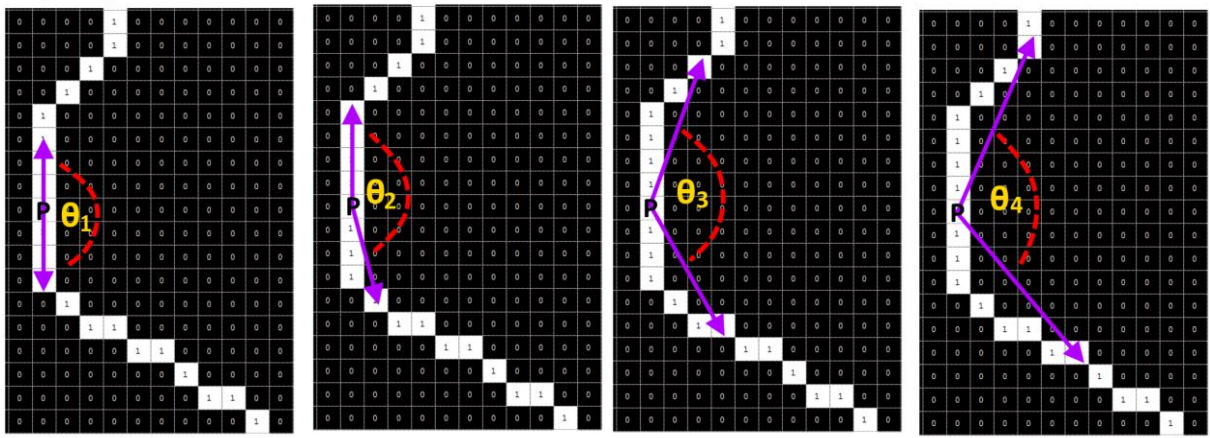

$$\theta(P) = \sum_{i=1}^{n} \frac{\theta_i}{n}$$

**Figure 2) a) The angle of the curvature (ϴ) at the point P of the vessel contour (ϴ (P)). b) The curvature angle at each point (P) of the vessel contour is different for different scales (ϴ$_1$, ϴ$_2$, ϴ$_3$….). c) The curvature angle at point P is taken as the average of the curvature angle at various scales: ϴ(P)=(ϴ$_1$+ϴ$_2$+ϴ$_3$…+ϴ$_n$)/n**

# 3. Applying a surface curvature map for improving vessel content recognition

Various of computer vision methods for the recognition of the boundaries of materials inside transparent vessels have been developed.[1,2,7,12,22] The main application of such methods is the recognition of properties such as fill level, liquid surface, and phase boundaries. These methods scan for a line or curve in the image that has the highest correlation with the boundary of the material in the vessel. This is done by looking for some image feature that acts as an indicator of the boundary of the material in the image and finding the curve with the highest correlation to this property in the image. Examples of such image features are edges and sharp intensity/colour changes. When evaluating the correspondence between a given curve and the material boundary in the image, each point of the curve receives a cost, or score, based on its correspondence to the image feature that act as an indicator for this boundary.[1,2] The cost or score of the entire curve is simply the average or percentile of all the points in the curve. The curve with the highest correlation to the material boundary indicator, in the image, receives the highest score or lowest cost.[1,2] Image features resulting from the vessel surface can disturb the process by changing the score/cost of the curves. Areas on the vessel surface with sharp curvature angles are likely to contain edges and sharp changes in intensity due to reflectivity change and functional parts.[36-39] It is desirable that curves passing through image region corresponding to areas with sharp curvature angle will have a lower chance to be used (higher cost/lower scores). To introduce the effect of surface curvature, the score of each point on a curve could be divided or multiplied by some factor ($F$) related to the curvature angle of the vessel surface at this point. A simple form of such a factor is: $F(P)=1+\theta(P)/C$. Where $F(P)$ is the curvature factor at point **P** and $\theta(P)$ is the vessel surface curvature angle at this point, **C** is a constant (for this work **C** was chosen as 30 degrees). Thus, if the curvature angle ($\theta(P)$) is zero (flat surface), then $F(P)=1$ and there is no effect on the score at this point (**P**). The higher the curvature angle ($\theta$), the higher the effect ($F$). If curve correspondence to the boundary of the material inside the vessel is evaluated by a cost function (the lower the cost the higher the match)[2] the price of each point in the curve is multiplied by the curvature factor ($F(P)$) at this point (**P**). If the correspondence to the boundary of the material inside the vessel is evaluated by a score (the higher the score the higher the match)[1] the score for each point (**P**) is divided by the curvature factor ($F(P)$) at this point. Thus, curves or lines which pass through areas with high vessel curvature will have their scores lowered or costs increased, and will be less likely to be identified as the material boundary in the vessel. Curvature factor maps for various of vessels are given in Figure 3.

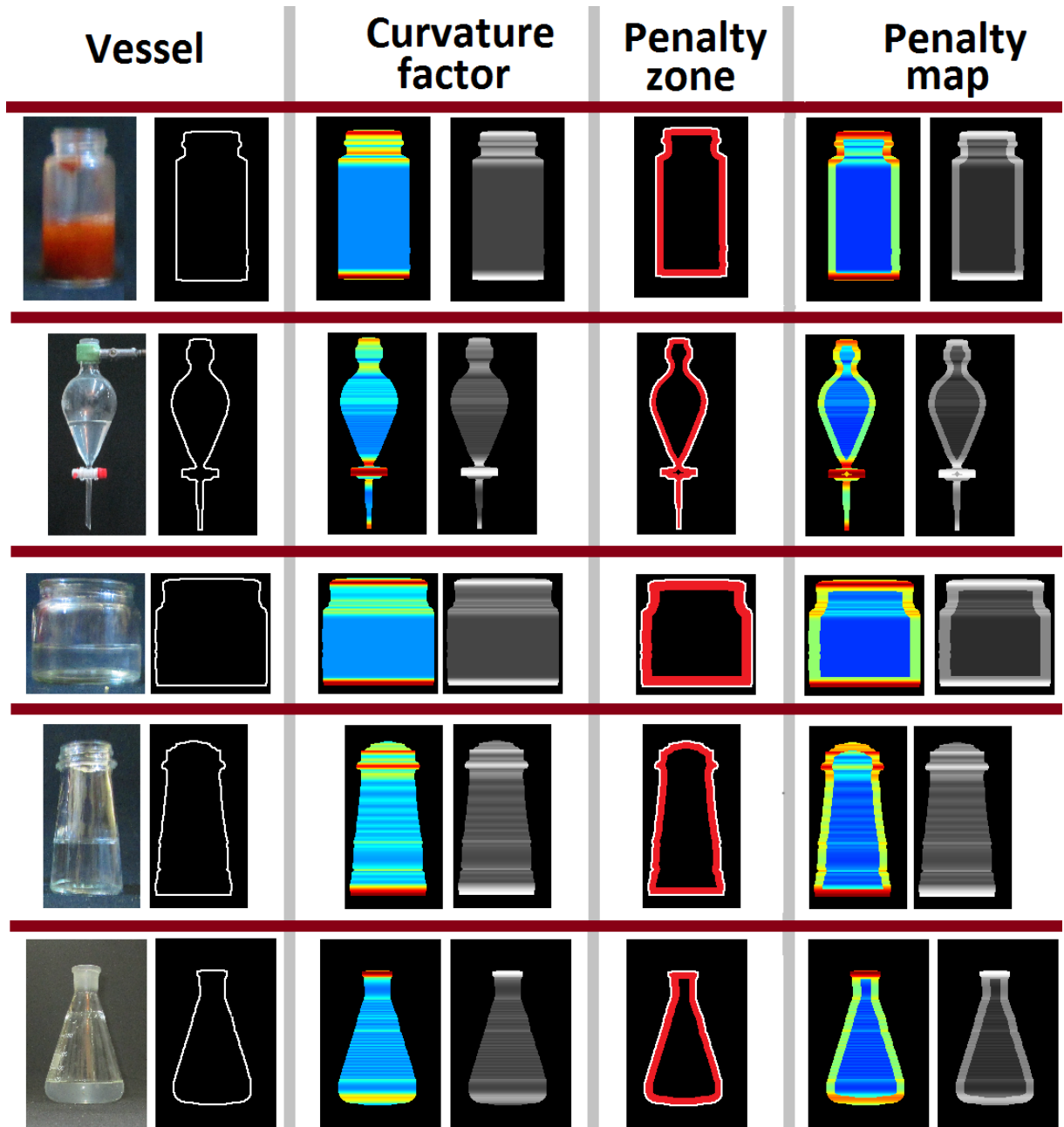

**Figure 3)** Curvature factor maps from various of vessels. **Vessel:** Image of the vessel and its is contour line in the image. **Curvature factor:** The curvature factor [$F(P)$] in each point of the image presented twice, once in colour (hot colours (red) correspond to high values), and once in grayscale bright regions represent high values. **Penalty zone:** The penalty zone around the vessel boundary marked red. **Penalty map:** Combined map of penalty zone and curvature factor give the penalty for boundary curve at each point of the image.

### 3.1. High cost zone near vessel boundary

Image areas located close to the boundary of the vessel in the image also tend to have strong edges that relate to the vessel and not the material inside it.[2] To account for this, the curvature factor ($F(\mathbf{P})$, Section 3) in image regions, close to the vessel boundary (Figure 3), was increased by two. For image regions which are a certain distance (10% of the vessel minimal dimension) from the vessel boundary in the image, $F$ was thus given as:

$F(\mathbf{P}) = 3 + \theta(\mathbf{P})/\mathbf{C}$. While for all other regions: $F(\mathbf{P}) = 1 + \theta(\mathbf{P})/\mathbf{C}$.

## 4. Method evaluation

### 4.1. Using the curvature factor to improve vessel content recognition

The method above can be combined with nearly any existing method for tracing material in transparent containers. To examine the ability of the method to improve recognition accuracy, the curvature factor ($F$, Section 3) was used with the method described in the paper "Tracing the boundaries of materials in transparent vessels using computer vision".[2] The method was applied twice, once with and once without the surface curvature factor. Unlike the original paper the scan for the boundary was applied to the entire region of the vessel. The results of each run are given in Tables 1-2 and Figure 4.

### 4.2. Evaluation

The recognition method was tested using a set of 231 images of transparent vessels with various materials. This set contained 130 images of vessels containing solid materials and 101 images of vessels containing fluids. The glass containers used in the images included ordinary glass vessels (e.g. jars, bottles and cups) as well as glassware used for analytical chemistry and organic synthesis (e.g. beakers, chromatographic columns, separatory funnels, Erlenmeyer flasks, round-bottom flasks, and vials).[41] The solids in the vessels included various powders, grained materials, and dry leaves with various particle sizes and morphologies; certain solids were immersed in liquids to examine the recognition of liquid solid interfaces. The liquids used in the images included water, oil, silica slurries, and various organic solvents (DMF, hexane, etc.). All pictures were taken using a uniform smooth black fabric with no folds as a tablecloth and background. The areas belonging to the vessel in the image were recognised using template matching or by extracting the vessel region from the uniform background based on its symmetry. The Matlab source codes for all methods and documentation are supplied in the supporting material.

### 4.3. Evaluating the test results

The material boundary inside the vessel was identified twice for each image: once with the curvature factor (*F*, Section 3) and once without. The material boundary found for each image by each method was manually compared to the boundary of the material in this image and assigned one of four levels of matching: 1) Full match: the path found and the material boundary in the image completely overlap; 2) Good match: the path found and the phase boundary in the image overlap nearly completely with minor deviations (approximately 90% or more); 3) Partial match: the path found overlaps with the material boundary in most areas, but deviates from it in major regions; 4) Low match: the path found mostly misses the material boundary in the image, but overlaps with it in a few regions; 5) Complete miss: the path does not overlap with the material interface in any region. For true/false evaluation of the results, only the perfect and good matches (Cases 1 and 2) were considered as true recognitions. Partial and low matches (Cases 3-5) were considered false recognitions. The results are given in Tables 1-2 and Figure 4.

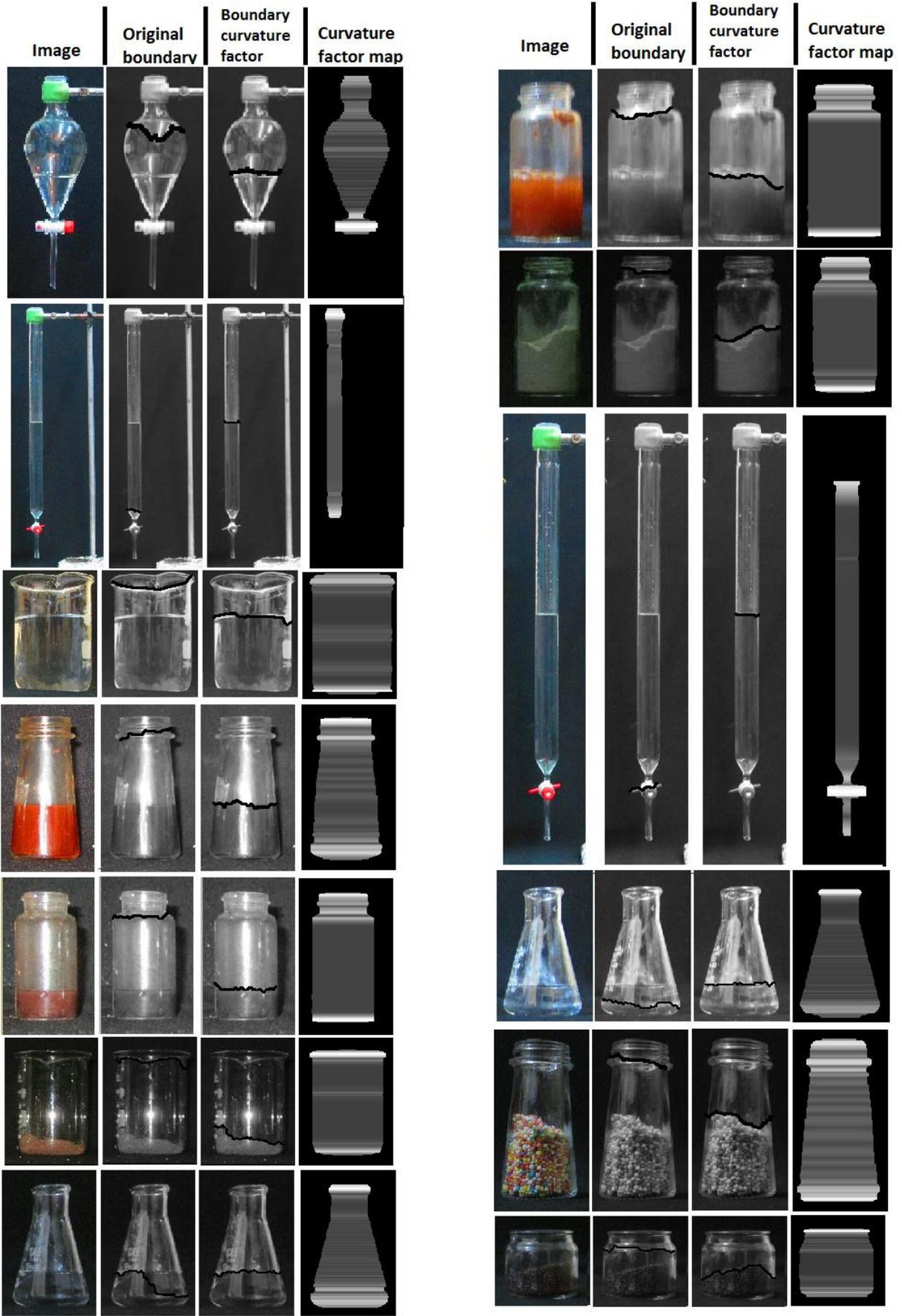

**Figure 4)** Recognition of materials interface in transparent vessels with and without curvature factor.

# 5. Results and discussion

The results (Figure 3) show that the method was able to identify vessel surface areas with sharp curvature angles as well as image areas corresponding to the functional parts of vessels, (corks and valves). Using the curvature factor ($F$, Section 3) enabled a considerable improvement in the recognition accuracy of materials interface inside the vessels. This can be seen by comparing the example of the recognition process with and without the use of curvature factor (Figure 4). As can be seen from Table 1, the fraction of false recognition of materials inside the vessel fell by 56% for liquid and by 25% for solids after using the curvature factor ($F$). The fraction of boundaries that were completely missed fell by over 66% for liquids and by over 50% for solids, after using the curvature factor. As can be seen in Figures 3-4, one of the main problems in deriving the curvature map is that the boundaries of the vessel in the image are often not accurately traced which leads to disturbances in the evaluation of the vessel curvature. Separatory funnels, vials, jars and chromatography columns show the highest reduction of miss rates after using the curvature factor. This is because these vessels contain highly curved surfaces and functional parts, such as corks and the valves. Vessels with uniform surface curvatures, such as cups and beakers, benefitted the least from using the curvature factor.

**Table 1: Recognition accuracy of liquid boundaries in transparent containers.**

| Liquids | True match[a] | False match[a] | Perfect match[a] | Small deviation[a] | Minority deviation[a] | Majority missed[a] | Complete missed[a] |
|---|---|---|---|---|---|---|---|
| Curvature factor[b] | 88% | 12% | 82% | 6% | 2% | 2% | 8% |
| No curvature[c] | 73% | 27% | 69% | 4% | 0% | 1% | 26% |

[a] See Section 4.3 for explanation of each field.
[b] Results from recognition when curvature factor used (Section 4.3).
[c] Recognition accuracy with no curvature factor (Section 4.3)

**Table 2: Recognition accuracy of solid material boundaries in transparent containers.**

| Solids | True match[a] | False match[a] | Perfect match[a] | Small deviation[a] | Minority deviation[a] | Majority missed[a] | Complete missed[a] |
|---|---|---|---|---|---|---|---|
| Curvature factor | 82% | 18% | 75% | 7% | 7% | 3% | 8% |
| No curvature | 76% | 24% | 72% | 4% | 5% | 0% | 19% |

[a] See Section 4.3 for explanation of each field.
[b] Results from recognition when curvature factor used (Section 4.3).
[c] Recognition accuracy with no curvature factor (Section 4.3)

# 6. Conclusion

This work proposed a general method for dealing with the main problem of machine vision-based recognition of objects and materials inside transparent vessels. This problem involves the ability to distinguish between image features resulting from vessel surfaces and features (edges), resulting from the materials inside the vessel interiors. This was solved by identifying the regions on the vessel surface with a high probability of showing strong edges. Two major assumptions regarding the vessel were made. The first assumption was that areas on the vessel surface that show sharp curvature angle are more likely to have strong edges due to reflectivity changes or functional parts (corks, valves). The second assumptions is that, given the high symmetry of most vessels (bottles, jars), it is possible to estimate the curvature of the vessel surface at a given point in the image by knowing the curvature of the vessel contour in the same row of the image. Thus, the angle of the curvature of the contour line of the vessel in the image is related to the angle of the vessel surface curvature in that row of the image. It was found that using these assumptions, it was not only possible to accurately recognise areas in the image corresponding to high curvature, but also to identify areas corresponding to a vessel's functional parts. Using this method to estimate the probability that a given edge in the image was cause by the vessel's surface allows considerable improvement in the automatic recognition accuracy of the contents of such vessels. The approach could be used with any computer vision method for the recognition of objects and materials in transparent containers and can enable considerable improvement in recognition accuracy.

# 7. Supporting Material

Source code for creating the curvature factor map for a given vessel is available at:

http://www.mathworks.com/matlabcentral/fileexchange/51028-create-a-curvature-factor-map-for--axisymmetric--vessels-based-on-their-outline-in-the-image

Source code for recognition of material boundaries in transparent vessels with use of the curvature factor is available at:

http://www.mathworks.com/matlabcentral/fileexchange/51029-find-the-boundaries-of-materials-in-transparent-vessels-using-computer-vision--curvature-adjustment-